\documentclass[letterpaper, 10 pt, conference]{ieeeconf}  
\IEEEoverridecommandlockouts                              
\pdfoutput=1

\usepackage{tabularray}
\usepackage{graphicx}
\usepackage{epstopdf}
\usepackage[utf8]{inputenc}
\usepackage{pgfplots}
\usepackage{amsmath,amssymb}
\usepackage{subfig}
\usepackage{xfrac}%for slant fraction
\usepackage{algorithm}
\usepackage[noend]{algpseudocode}
\usepackage{bm}
\usepackage{wrapfig}
\usepackage{float}
\usepackage{orcidlink}
\usepackage{mathtools}
\usepackage{comment}
\usepackage[super]{nth}
\floatstyle{plaintop}
\restylefloat{table}

\usepackage{cite}

\pgfplotsset{compat=1.17}

\usepackage{xcolor}

\title{Pursuit-Evasion for Car-like Robots with Sensor Constraints}

\author{Burak M. Gonultas$^{1}$\orcidlink{0000-0002-7966-7929} and Volkan Isler$^{1}$\orcidlink{0000-0002-0868-5441}% <-this % 
\thanks{$^{1}$ the authors are with the Department of Computer Science and Engineering, University of Minnesota, Minneapolis, MN, 55455, USA
        {\tt\small \{gonul004, isler\}@umn.edu}}
        \thanks{This work was supported by the MN LCCMR program.}%
}

\usepackage{fancyhdr}
\fancypagestyle{firststyle}
{
\fancyhf{} 
}
\begin{document}
\maketitle
\thispagestyle{firststyle}
\begin{abstract}
    We study a pursuit-evasion game between two players with car-like dynamics and sensing limitations by formalizing it as a partially observable stochastic zero-sum game. The partial observability caused by the sensing constraints is particularly challenging. As an example, in a situation where the agents have no visibility of each other, they would need to extract information from their sensor coverage history to reason about potential locations of their opponents. However, keeping historical information greatly increases the size of the state space. To mitigate the challenges encountered with such partially observable problems, we develop a new learning-based method that encodes historical information to a belief state and uses it to generate agent actions.
    %We develop a novel belief network, named \textit{BiMDN} for extending multi-agent deterministic policy gradient algorithms. Using BiMDN, we simultaneously obtain strategies for both players. 
    Through experiments we show that the learned strategies improve over existing multi-agent RL baselines by up to 16\% in terms of capture rate for the pursuer. %while the learned evader model demonstrates up to 5\% better escape rate over the heuristic baselines even against our competitive multi-agent RL pursuer model. 
    Additionally, we present experimental results showing that learned belief states are strong state estimators for extending existing game theory solvers and demonstrate our method's competitiveness for problems where existing fully observable game theory solvers are computationally feasible. Finally, we deploy the learned policies on physical robots for a game between the F1TENTH and JetRacer platforms moving as fast as $\textbf{2\ m/s}$ in indoor environments, showing that they can be executed on real-robots. % Our code and supplementary material including videos from experiments are available at \url{https://gonultasbu.github.io/pursuit-evasion/}.
    %We also present experiment results which show how the pursuit-evasion game and its results evolve as the player dynamics and sensor constraints are varied.
\end{abstract}
\section{Introduction}

Pursuit-evasion games are optimization problems in which one or more pursuers try to ``capture" an evader.
The notion of capture is usually formulated as being co-located or getting close. However there are other formulations such as visibility-based pursuit evasion~\cite{guibas1999visibility,isler2005randomized} where capture is achieved when the evader is located. In robotics, pursuit-evasion games are studied to model motion planning problems in dynamic scenarios and to investigate the coupling between player dynamics, observation models and environment complexity. 

As an example, in the classical homicidal chauffeur game~\cite{isaacs1999differential}, the pursuer is a car-like vehicle whose goal is to hit a pedestrian who is not subject to any turning constraints. In this game, the pursuer has three state variables (position and orientation) and the evader has two (only position). One might view the solution of this game as a subspace (manifold) of this five dimensional joint state space. The tangent space of this solution manifold corresponds to the optimal control actions of the players. 
In some cases, the differential equations for the evolution of the game when the players play optimally can be derived in closed form. The strategies for a given instance are usually obtained by numerical integration. Alternatively, the state space can be discretized and the solution can be obtained using dynamic programming. 

\begin{figure}[!t] 
\centering
\subfloat{\includegraphics[width = 1.00\linewidth]{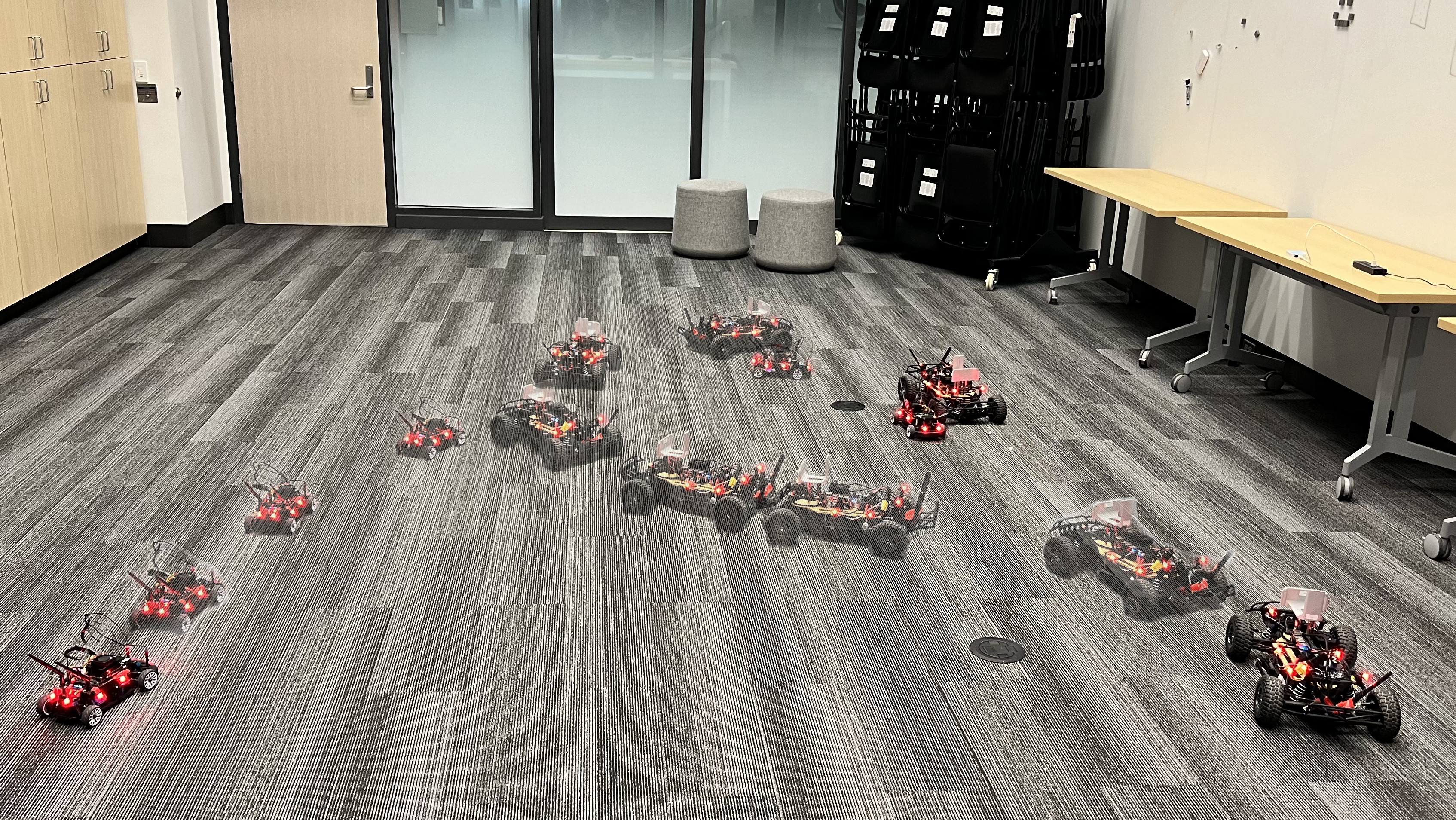}}
\caption{\textbf{Real-world experiments:} Execution of the learned pursuit-evasion policies on two robotic cars with different wheelbase, velocity and steering angle constraints. %We train both policies for the pursuer and evader against each other using multi-agent reinforcement learning.
}
\vspace{-20pt}
\label{fig:init_dist}
\end{figure}

However, such methods become quickly intractable as the dimensionality of the game increases. For example, in his book, Isaacs presents only a partial solution for the ``game of two cars" where both players have car-like dynamics. The dimensionality issue becomes more severe when sensing is involved; in the standard formulations, it is assumed that the players can observe each others' state fully at each time step. This may not be always true due to observability limitations. The resulting ``partial observability games" have even higher (sometimes infinite) dimensionality as the players now need to reason about all feasible states that are consistent with their observation histories.  

\begin{figure}[t]
\centering
\subfloat{\includegraphics[width = 0.98\linewidth]{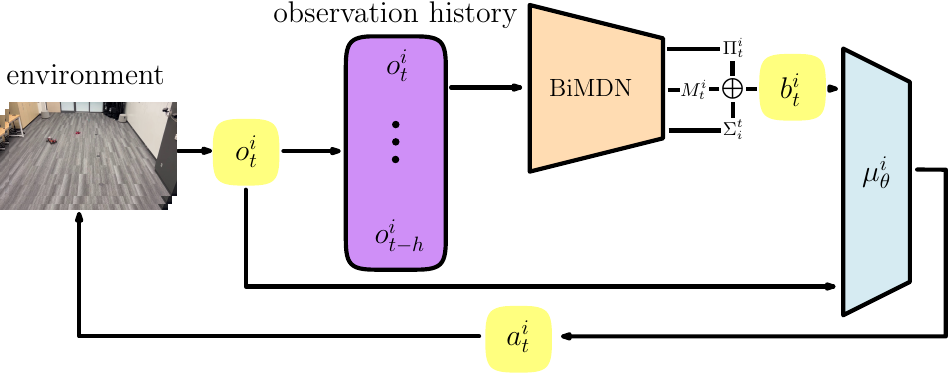}}
\caption{\textbf{Our method:} Agents receive partial observations $o^i_t$ at each timestep and store observation memories. For each agent, our BiMDN models encode individual agent observation histories to belief states $b_t^i$, whose uncertainty can be explicitly quantified. Trained actor networks $\mu_\theta^i$ use combined information from the belief states and environment partial observations to compute actions $a^i_t$.}
\vspace{-15pt}
\label{fig:bimdn}
\end{figure}

Reinforcement Learning (RL) provides an alternative approach to construct the solution manifold by learning it through game play.
%Modern RL methods often represent the solution as a neural network which takes the observations as input and returns a distribution over the actions. 
In this paper, we extend recently developed multi-agent reinforcement learning (MARL) techniques by combining Mixture Density Networks \cite{bishop1994mixture} and bidirectional Long Short-term Memory (LSTM) Networks \cite{hochreiter1997long, schuster1999better, bazzani2016recurrent} to build a novel belief network as shown in Fig. \ref{fig:bimdn}, named \textit{BiMDN}. We demonstrate that our proposed method can solve a challenging game where the pursuer is a car with five state variables including velocities better than existing MARL algorithms by encoding agent observation histories to belief states. We investigate two versions of the evader. It is either a car with the same kinematic model as the pursuer or a point-mass. Hence, the joint-state dimensionality can be as high as 10. The players are subject to field-of-view sensing restrictions for both the viewing angle and viewing distance (Section~\ref{Problem Formulation}).
 We cast the game as a competitive zero-sum Partially Observable Stochastic Game (POSG) and use the MARL algorithm \emph{multi-agent deep deterministic policy gradient (MADDPG)}~\cite{lowe2017multi} with BiMDN to simultaneously train pursuer and evader policies. Our problem formulation and proposed approach do not require any pretrained or designed expert evader policy and generalize over varying agent constraints.
 
% We further show the practical applicability of the resulting strategies by implementing a testbed with ...\\
 
Our contributions are as follows:
%  \begin{itemize}
%     \item Using a multi-agent reinforcement learning approach we train pursuer and evader policies for a pursuit evasion game with dynamic and sensor constraints.
%     \item We show that our method is able to train competitive pursuer and evader policies without the need for expert policies. In our experiments we individually compare pursuer and evader policies against existing baselines as well as compare them against each other.
%     \item We present how the pursuit-evasion game and its results evolve as the player dynamic and sensor constraints are varied.
%     \item We deploy learned policies on physical robots for a game between the F1TENTH\cite{o2020f1tenth} and JetRacer \cite{jetracer} platforms; showcasing the experimental results of a successful policy transfer to real world.
% \end{itemize}

 \begin{itemize}
    \item We introduce a novel belief network named \textit{BiMDN} for MARL to solve a pursuit-evasion game with complex dynamics and sensing constraints. %BiMDN is a bidirectional recurrent mixture model whose uncertainty can be explicitly quantified. 
    We show that our method is able to train stronger pursuer policies without the need for expert policies even as compared to vanilla MARL algorithms. 
    %\item We provide insights on how the pursuit-evasion game and its results evolve as the player dynamic and sensor constraints are varied.
    \item In simulation experiments we compare our pursuer and evader policies to differential game solver baselines from the reachability literature and demonstrate competitive performance for simpler problems where existing solvers are computationally feasible.
    \item We deploy the learned policies on physical robots for a game between the F1TENTH\cite{o2020f1tenth} and JetRacer \cite{jetracer} platforms moving at high-speed and showcase a successful direct policy transfer to real world.
\end{itemize}

%\vtxt{here we need to say something about what we found: do we think that these methods can sufficiently solve the game for all conditions; how do we know this?} 

Overall, our results show the potential of recurrent mixture models coupled with multi-agent reinforcement learning in navigating complex pursuit-evasion scenarios within dynamic environments and demonstrate that simulation-trained policies are directly transferable to real autonomous agents.

% Source code is available on our project page: \href{will be released after review}{\color{magenta}{will be released after review}}

\section{Related Work}\label{Related Work}
%In this section, we will cover previous research that is most relevant to our proposed approach. Since our method is inspired by the original pursuit evasion problem, we will first go over some of the publications formulating various extensions of the original pursuit-evasion problem. After that, we will discuss previous publications proposing various Multi-Agent Reinforcement Learning (MARL) methods and their applications on dynamic games.
In this section, we review relevant pursuit-evasion and MARL literature. 
\vspace{-10pt}
\subsection{Pursuit-Evasion}

The mathematical origins of the pursuit-evasion problem can be traced back to 1930's as the Lion and Man problem. In the original formulation, a lion and a human are in a circular area where the lion tries to catch an escaping human as quickly as possible.
%when the human tries to avoid the lion for as long as possible. 
%Since then, 
Various variants of this game have been studied in the robotics literature~\cite{chung2011search}. In his seminal work \cite{isaacs1999differential} Isaacs presented a differential game based formulation to the pursuit-evasion problem which has served as the basis of game-theoretic approaches. % to the pursuit-evasion problem\cite{bhattacharya2011cell}.

%have been formulated and studied in different data structures such as trees \cite{kolling2009pursuit}, graphs \cite{parsons2006pursuit}, polygons \cite{guibas1999visibility,isler2005randomized} and surface of polyhedra \cite{noori2014lion}. 

In control-theoretic formulations, both of the agents in a pursuit-evasion problem have constraints on maximum velocities. One prominent example is the Homicidal chauffeur problem formulated by Isaacs \cite{isaacs1999differential}, which is a special case of the pursuit-evasion problem where the pursuer has nonholonomic constraints. Since then, numerous works have studied pursuit-evasion games with dynamic constraints\cite{exarchos2015suicidal, ruiz2016differential, scott2018optimal}. In \cite{de2021decentralized} the authors propose a single-agent reinforcement learning approach to the multi-pursuer with unicycle constraints scenario against a single, omnidirectional evader. Similarly, in \cite{yang2023large} Yang et al. study a multi-pursuer, multi-evader version of pursuit-evasion problem with unicycle constraints.% and propose a MARL state processing method to improve the scalability.

Reachability literature presents solutions to various pursuit-evasion problems. ToolboxLS \cite{mitchell2008flexible} demonstrates an Hamilton-Jacobi (HJ) solver by formulating pursuit-evasion as a differential game. HelperOC \cite{helperOC} and OptimizedDP \cite{bui2022optimizeddp} establish further improvements in performance and handling a higher number of dimensions for HJ solvers. Similarly, game-theoretic studies also treat pursuit-evasion games as general-sum games, provide various solvers \cite{fridovich2020efficient, le2022algames, peters2022learning, liu2023learning, zhu2023sequential, schwarting2021stochastic} searching for Nash equilibria.

If the players can not observe each other at all times, %for example, if the game is played in  a complex environment under sensing limitations, 
pursuit-evasion game has been defined as a combination of two subcomponents: 1) locating the evader (search problem) and 2) capturing the evader. From the pursuer perspective, the search problem appears once sensor constraints are introduced to the pursuit-evasion problem. From the evader perspective, optimal strategies depend on the environment as well as the sensor constraints of the agents in the pursuit-evasion game. The problem of planning paths to reach a state with unobstructed view of the evader was first presented in \cite{suzuki1992searching}. In \cite{guibas1999visibility}, it was shown that in a simply connected polygon of $n$ vertices, $\Theta(\log{n})$ pursuers are necessary and sufficient to detect an unpredictable evader. Numerous studies since then have formulated different variations of the pursuit-evasion problem, \cite{gerkey2006visibility, stiffler2017persistent,shishika2018local} with \cite{gerkey2006visibility} studying the case with a pursuer with limited field of view and \cite{stiffler2017persistent} with unreliable sensor data.
%Some of the studies working in a scenario with sensor constraints are as follows: 
In \cite{isler2005randomized} Isler et al. propose a randomized algorithm to locate the evader with a single pursuer and extend this strategy to two pursuers with line-of-sight visibility to capture the evader in simply connected polygons. In later work, \cite{noori2014lion} Noori and Isler have shown that a single pursuer can locate and capture the evader in monotone polygons. In \cite{engin2021learning} Engin et al. have proposed a single-agent reinforcement learning method with a compact belief representation for pursuit-evasion games with visibility constraints in simply-connected polygons. In \cite{zhang2022game} Zhang et al. propose a MARL-based approach for n-pursuer pursuit-evasion game between drones in urban maps, incorporating an LSTM network for future state predictions in simulated experiments. In \cite{bajcsy2023learning} Bajcsy et al. have studied the pursuit-evasion problem for quadrupedal robots under real-world physical and visibility constraints.% and proposed a teacher-student network pair to deal with %the issues caused by 
%partial observability. %in the problem definition.

%The players in these games have simple point-robot dynamics.
In this work, we differentiate ourselves from existing literature by studying a game between players with complex dynamics \emph{and} sensing limitations; assuming no prior knowledge of the dynamics or the sensor model, which are typical limitations of the methods proposed in the reachability and game theory solver literature. Additionally, unlike some of the prior works, we do not train against an expert policy to showcase competitive performance in simulation as well as real robot experiments.
\vspace{-10pt}
\subsection{Dynamic Games \& MARL} 
Isaacs' \cite{isaacs1999differential} and other game-theoretic formulations \cite{shapley1953stochastic, littman1994markov} have long influenced robotics, controls and reinforcement learning research \cite{mitchell2005time,zhang2021multi,pinto2017robust,bucsoniu2010multi}. Significant progress has been made using MARL algorithms \cite{lowe2017multi, ackermann2019reducing, de2020independent,rashid2020monotonic} in settings where large-scale simulations are feasible such as Starcraft, Dota 2, Go and Chess \cite{vinyals2019grandmaster,berner2019dota,silver2018general,silver2016mastering}. Although there are recent works demonstrating successful reinforcement learning methods' deployment on physical systems \cite{bajcsy2023learning, brunnbauer2022latent}, joint learning in game settings and deployment on physical systems still rema\vspace{-10pt}ins a challenge for the robotics community.
\vspace{-10pt}
\subsection{Belief States}
A significant number of problems in robotics domain inherently present some form of partial observability which breaks Markovian assumption made by standard RL literature. %While it is possible to feed the full observation history to existing algorithms, this approach quickly becomes infeasible once the state space moves beyond toy problems. 
There is a body of research proposing the use of belief states \cite{monahan1982state, kaelbling1998planning, silver2010monte}, which are observations conditioned on the history of interactions to preserve Markovian properties. Contemporary partially observable RL literature presents various recurrent models for belief modeling \cite{wayne2018unsupervised,han2019variational,moreno2021neural,wang2023learning}, aiming to capture spatiotemporal features relevant to the downstream task. With BiMDN, we additionally focus on modeling the spatial uncertainty presented by the pursuit-evasion problem to learn stronger policies in multi-agent problems.

\section{Problem Formulation}\label{Problem Formulation}
In this section, we introduce the relevant terminology and models used in our problem definition.
\vspace{-10pt}
\subsection{Agent and Environment Models}\label{Models}
\begin{figure}[]
\centering
\subfloat[]{\includegraphics[width = 0.49\linewidth]{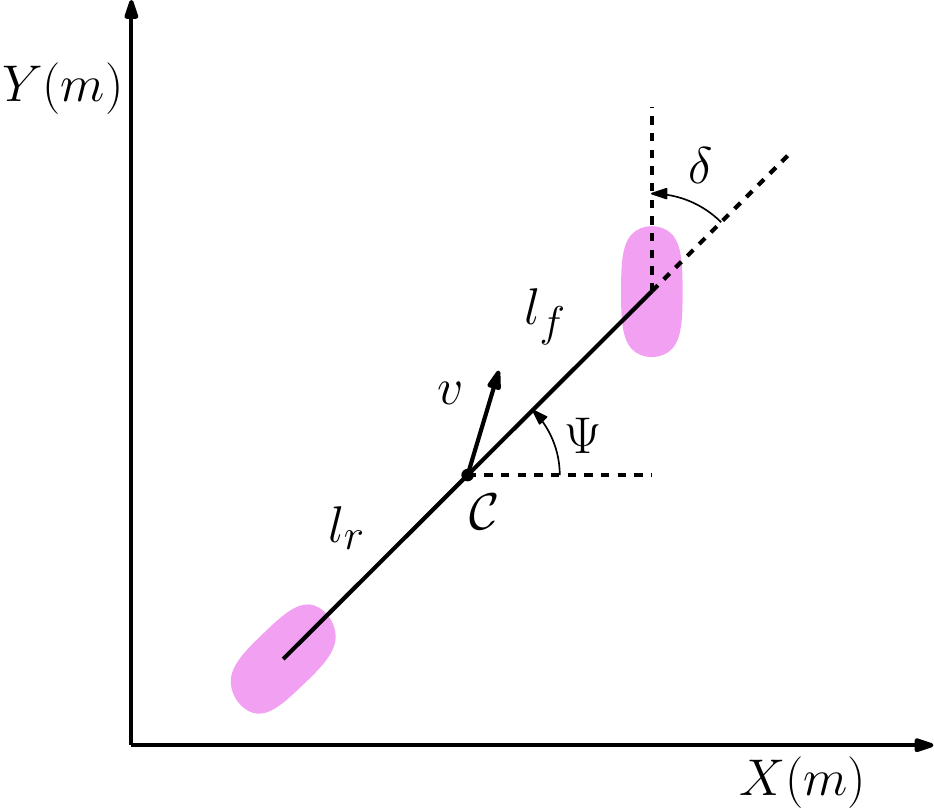}\label{fig:model_car}}
\subfloat[]{\includegraphics[width = 0.49\linewidth]{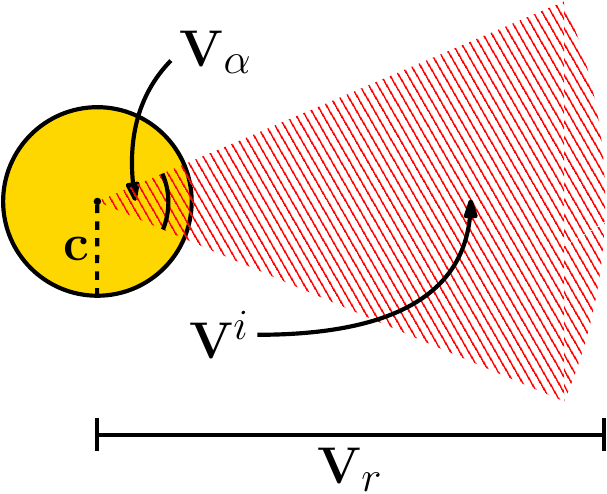}\label{fig:model_fov}}

\caption{\textbf{Car kinematic bicycle model and sensor representations:}  For simulated experiments, the pursuer uses the kinematic bicycle model presented in (a). For physical experiments, both agents use the kinematic bicycle model with different parameters. All agents use the same sensor model presented in (b).  }
\vspace{-15pt}
\label{fig:vehicle_ipes}
\end{figure}
The pursuer kinematic model, which is represented by the state-space model~\cite{althoff2017commonroad}, is given in Eq~\eqref{eq:adp_vdot}. It consists of the following states $x_1=s_x, x_2=s_y,x_3=~\delta, x_4=v, x_5= \Psi$.
\begin{equation}\label{eq:adp_vdot}
\begin{aligned}
&\dot{x_1} = x_4 \cos (x_5) && \dot{x_2} = x_4 \sin (x_5)\\
&\dot{x_3} = f_{\mathrm{steer}}(x_3,u_1) && \dot{x_4} = f_{\mathrm{acc}}(x_4,u_2)\\
&\dot{x_5} = \frac{x_4}{l_{f}+l_{r}} \tan (x_3)\\
\end{aligned}    
\end{equation}
where $s_x$ and $s_y$ are the coordinates in meters in the global frame.
%where $s_x$ is the $x$ position in global frame of reference in meters, $s_y$ is the $y$ position in global frame of reference in meters, 
$v$ is the longitudinal velocity in $m/s$, and $\Psi$ is the global yaw angle at the vehicle center in radians. Inputs are defined as $u_{1}$ representing the steering velocity $\dot{\delta}$ in $rad/s$ and $u_{2}$ representing the longitudinal acceleration $\dot{v}$ in $m/s^2$. System steering and acceleration input constraints are represented as $f_{steer}$ and $f_{acc}$, respectively. $l_{f}+l_{r}$ is the car model parameter representing the vehicle wheelbase with $l_{f}$ representing the distance of the front axle from the center of gravity and $l_{r}$ representing the same for the rear axle in meters with zero slip angle assumption. See Fig. \ref{fig:model_car}.

The evader is based on the point-mass model  \cite{althoff2017commonroad} defined in Eq. \eqref{eq:pm_vdot}, consisting of the following states $x_1 = s_x, x_2 = s_y, x_3 = \dot{s_{x}}, x_4 = \dot{s_{y}}$ and uses the same global frame of reference notation with the inputs $u_{1}$ and $u_{2}$ representing the accelerations along the global x and y-axes $\ddot{s_x}$ and $\ddot{s_y}$ in $m/s^2$. 

\begin{equation}\label{eq:pm_vdot}
\dot{x_1} = x_3, \quad \dot{x_2} = x_4, \quad \dot{x_3} = u_1, \quad \dot{x_4} = u_2
\end{equation}

We denote the rectangular workspace by $\mathcal{E}$ as a subset of the 2D Cartesian coordinate system defined by the minimum and maximum values along each axis such that $\{X_{low}, X_{high}, Y_{low}, Y_{high}\} \subset \mathbb{R}$. $\mathcal{E}$ is assumed to be obstacle free and fully traversable. Euclidean distance between two points in $\mathcal{E}$ of arbitrary size is denoted by $d(z_{1},z_{2})$ with $z_1,z_2 \in \mathbb{R}^2$. All distances are defined in meters.

Pursuer agent uses the noise-free sensor model represented by a wedge (slice of a circle) centered at agent center of mass and have two parameters: field of view angle ($rad$) $V^i_{\alpha}$ and range ($m$) $V^i_r$ as shown in Fig. \ref{fig:model_fov}. An agent sees its opponent if and only if the opponent is inside the agent sensor footprint. The sensor footprint of an agent i is denoted by $V^i \subseteq \mathcal{E}$ with $i \in \{p,e\}$. Therefore, the pursuer $p$ sees the evader $e$ if $e \in V^p$ and vice versa. The sensor footprint of an agent is presented in Fig. \ref{fig:model_fov} and the visibility condition is also defined as follows with the help of a binary visibility variable $\mathbb{I}$:
$$
\mathbb{I}^i =
\begin{cases}
    +1, & \text{if } d\bigl((x_1^i,x_2^i),(x_1^j,x_2^j)\bigr) \leq V^i_r \text{ and } \\
       & \quad \frac{-V^i_{\alpha}}{2} \leq \operatorname{atan2}(x_2^j-x_2^i,x_1^j-x_1^i)-x^i_5 \leq \frac{V^i_{\alpha}}{2}, \\
    -1, & \text{otherwise}.
\end{cases}
$$
where $i$ and $j$ denote an agent and its opponent, respectively. Angular differences are wrapped to the interval $[-\pi,\pi)$. Measurement function is defined formally as follows :

\begin{equation}\label{eq:sensors-formal}
\mathbb{M}^i =
\begin{cases} 
\mathbf{x}^j, & \text{if } \mathbb{I}^i = +1, \\
\mathbf{0} \in \mathbb{R}^n, & \text{otherwise}.
\end{cases}
\end{equation}
where the function returns a vector of opponent states or zeros depending on the observability. Since point-mass agents don't have a heading, $V^i_{\alpha}$ is always $2 \pi\ rad$ for point-mass agents, making the angular constraint negligible and forming a disk with the footprint.

%as defined in Eq. \ref{eq:sensors}
%\begin{equation}\label{eq:sensors}
%\begin{aligned}
%a = 
%\begin{cases}
%    2x + 1, & \text{if } x > 0 \\
%    x^2 - 1, & \text{if } x \leq 0
%\end{cases}
%\end{aligned}    
%\end{equation}

%\subsection{Metric Workspace}\label{Workspace}

%We consider a global frame of reference such that the four corners of the rectangular workspace are located at $(X_{low},Y_{low})$, $(X_{high},Y_{low})$, $(X_{high},Y_{high})$, $(X_{low},Y_{high})$ in meters with $X_{low}, X_{high}, Y_{low}, Y_{high} \in \mathbb{R}$
\subsection{Game Formulation}

Within the scope of this paper, the pursuit evasion game is defined as a POSG. However, it must be noted that without added stochastic noise, presented agent dynamics are fully deterministic as described in Eq. \ref{eq:adp_vdot} and Eq. \ref{eq:pm_vdot}. 

%The formal definition of a POSG is shown in \ref{def:posg} as defined by Terry et al. \cite{terry2021pettingzoo} inspired by Shapley \cite{shapley1953stochastic}.

%\begin{definition}\label{def:posg}
%    A Partially Observable Stochastic Game (POSG) is a tuple $ \langle \mathcal{S}, s_0, N, A^{i},P,R^{i},\Omega^{i},O^{i} \rangle$, where:
%
%    \begin{itemize}
%    \item $\mathcal{S}$ denoting the set of possible states.
%    \item $s_{0}$ denoting the initial state.
%    \item $N$ is the number of agents. The set of agents is denoted as $[N]$
%    \item $\mathcal{A}^{i}$ is the set of possible actions for agent $i$ with $i \in [N]$.
%    \item $P:\mathcal{S}\times\prod_{i}\mathcal{A}^{i} \times\mathcal{S}$$\rightarrow[0,1]$ is the transition function. It has the property that for all $s\in\mathcal{S}$, for all $(a^1,a^2,\ldots,a^N) \in \prod_{i}\mathcal{A}^{i}$,$\sum_{s' \in \mathcal{S}}P(s, a^1, a^2, \ldots, a^N, s') = 1.$
%    \item $R^i:\mathcal{S}\times\prod_{i}\mathcal{A}^{i}\times\mathcal{S}\rightarrow \mathbb{R}$ is the reward function for agent $i$.
%    \item $\Omega^i$ is the set of possible observations for agent i.
%    \item $O^i:\mathcal{A}^{i}\times\mathcal{S}\times\Omega^i\rightarrow[0,1]$ is the observation function. It has property that $\sum_{\omega\in\Omega^{i}}O^{i}(a,s,\omega)=1$ for all $a \in \mathcal{A}^{i}$ and $s \in \mathcal{S}$
%    \end{itemize}
%\end{definition}

We are given a rectangular workspace $\mathcal{E}$. The positions of the agents at timestep $t$ are defined as $p_t$ and $e_t$ for the pursuer and the evader, respectively. At each timestep both agents make their moves simultaneously. The pursuer captures the evader if $d(p_t, e_t) \leq 2c$ with $c$ being each agent's radius. The pursuer wins the game if a capture occurs within an allocated timestep limit $\mathcal{T}$. Timeouts are considered evader victories.

\section{Method}\label{Proposed Method}
The objective of each agent is to maximize its total expected reward $G^i = \sum_{t=0}^{\mathcal{T}}{\gamma_{t}r^i_{t}}$ over time horizon $\mathcal{T}$ where $r^i_t$ is provided by the environment at timestep $t$ and $\gamma \in [0,1]$ is the discount factor. Per-agent actions are sampled from the agent policy networks $a^{i}_{t}\sim\mu^{i}_{\theta}(s^{i}_{t})$.
\subsection{Multi-Agent Deep Reinforcement Learning}
We use the MADDPG algorithm \cite{lowe2017multi} which follows the Centralized Critic, Decentralized Execution (CTDE) principle to train the agents by making them play against each other from scratch with access to privileged information on the opponents during training. This is different from self-play, where agents are completely independent during the training stage without access to privileged information. Our method is also model-free, i.e. agents are not provided with information on the dynamic equations that govern the interactions or learn such equations. Instead, both agents train corresponding value and policy networks through interactions.
\vspace{-5pt}
\subsection{Space Representations}
The hidden full state of the pursuit-evasion game between a kinematic car-like pursuer and a point-mass evader is referred as a 9-vector $s \coloneqq [x^p_1, x^p_2, x^p_3, x^p_4, x^p_5, x^e_1, x^e_2,x^e_3,x^e_4]^{\mathsf{T}}$. Without loss of generality, the state vector comprises of state variables presented in Section \ref{Models}. Time indices are dropped for brevity. Each agent in the pursuit-evasion game creates its own observation vector $o^i$ through partial observations from the environment represented as 11-vectors of floating point numbers by appending a binary observation flag $\mathbb{I}^i$ indicating the visibility of the opponent and a time index. If an agent cannot observe its opponent at a timestep, corresponding values in $o^i$ are zeroed and the binary flag $\mathbb{I}^i$ is set to $-1$. Observation vectors are normalized to contain values in interval $[-1,1]$. Input/action representations follow the same convention described in \ref{Models} and are represented as 2-vectors normalized to the interval $[-1,1]$ by taking acceleration, velocity, steering angle velocity and steering angle limits. Both the state and the action spaces are continuous.
\vspace{-5pt}
\subsection{Belief Network}
We extend each agent's observation vector with information derived from belief states $b^i_t$. BiMDN is a sequential model of bidirectional LSTM layers followed by fully connected layers and output heads for predicting a Mixture of Gaussians parameterized by the means $M^i_t$, standard deviations $\Sigma^i_t$ and cluster weights $\Pi^i_t$ using past and current observations. Predictions are made over 2 variables which represent the agent beliefs along the $x$ and $y$ axes. Prediction clusters are axis-aligned ellipsoids which is equivalent to a diagonal covariance matrix as we have experienced no significant improvement in performance by further parametrization of the covariance matrix. Similar to the CTDE paradigm (or asymmetric RL), training of BiMDN uses privileged information as ground truth with Negative Log Likelihood loss derived from $M^i_t$, $\Sigma^i_t$ and $\Pi^i_t$. %The agents use weighted means estimates from the belief networks' \textit{iff} opponent is not observable during inference.  

% are flattened to vectors and concatenated to the observations after normalization.

BiMDN learns to encode relevant aspects from a time series input of stacked historical observation vectors to a compact belief state. During training, unlike commonly used neural network architectures like LSTM networks, our architecture explicity quantifies uncertainty about its belief and does not output point estimates. This is a useful property in pursuit-evasion games where an average point estimate would be incapable of modeling the potential regions the opponent may be at. To indicate a higher uncertainty, the clusters expand in size proportional to their standard deviations and get dispersed across the environment. %If the opponent is observed recently, the clusters collapse to a single point. 
Further discussion including network parameters is provided in Section \ref{Multi-Agent Training}.% and depicted in Fig. \ref{fig:bimdn-details}.
\vspace{-5pt}
\subsection{Reward Structure}
We formulate pursuit-evasion problem as a competitive zero-sum POSG. Players have opposite rewards at each timestep, that is $r^p_t + r^e_t = 0$. At each timestep, the pursuer receives the following reward:
$$
r^p_t=\begin{cases}
            -\lambda_{capture}, & \text{if $t \geq \mathcal{T}$}\\
            \lambda_{capture}, & \text{if $d \leq 2c$ and $t < \mathcal{T}$}\\
            -(\lambda_t + \lambda_{d}\cdot d), & \text{otherwise}
		 \end{cases}
$$

\noindent where $\lambda_{capture}$, $\lambda_{t}$, $\lambda_{d}$  represent the coefficients for capture reward, per-timestep reward and distance reward, respectively. Each episode has two independent reset conditions, the capture condition (pursuer wins) or the timeout (evader wins). $d$ represents the Euclidean distance between agents as described in Section \ref{Problem Formulation}. Per-timestep rewards encourage the pursuer to capture as quickly as possible while distance rewards provide a denser reward structure to help the training procedure.

\begin{figure}[]
\centering
\subfloat{\includegraphics[width = 0.8\linewidth]{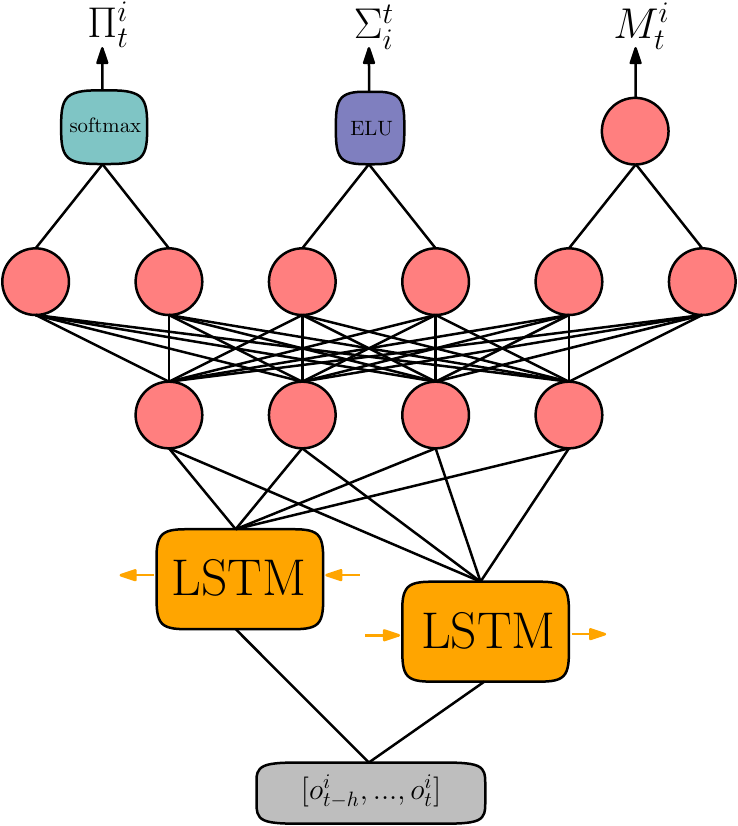}}
\caption{\textbf{BiMDN architecture:} BiMDN uses 1 layer of bidirectional LSTM, %which contain 2 seperate LSTMs for forward and backward passes 
with 128 hidden units followed by a fully connected layer of 32 output features. The network thereafter splits into 3 heads with a fully connected layer in each head before the output layers. %For cluster weights $\Pi^i_t$, the number of output features is equal to the number of cluster centers in the corresponding layer, other heads have twice the number of output features, representing 2D coordinates.
Outputs' activation functions are softmax, Exponential Linear Unit (offset with +1 to enforce nonnegativity) for cluster weights $\Pi^i_t$ and standard deviations $\Sigma^i_t$, respectively. Gaussian cluster centers $M^i_t$ are directly predicted by fully connected layers. Fully connected layers are represented in red (not the actual number of parameters) and number of output clusters is empirically set at 3.}
\vspace{-15pt}
\label{fig:bimdn-details}
\end{figure}

\section{Simulation Experiments}\label{Simulation Results}
In this section, we analyze our method by evaluating it against pursuer and evader baselines in a series of experiments. We have designed our experiments to answer the following questions: \textbf{Question 1:} How well do trained pursuer and evader agents perform against baseline methods? \textbf{Question 2:} Does BiMDN improve model performance over existing MARL methods? %\textbf{Question 3:} Qualitatively, are there any interesting emerging behaviors? %\btxt{\textbf{Question 4:} How do the model parameters affect agents' performances?}

% \begin{figure*}[!htb] 
% \centering
% \subfloat{\includegraphics[width = 1.0\linewidth]{figures/circular_multi_plot.pdf}}
% \caption{\textbf{250 initial distributions sampled uniformly at random:} Initial positions are transformed with respect to the pursuer frame of reference. Colors indicate the normalized time-to-capture. Evader starting points that are farther away from pursuer have higher Time-to-Capture.}
% \label{fig:init_dist}
% \end{figure*}

As the heuristic pursuer baseline, we use a modified pure pursuit controller \cite{coulter1992implementation}: if the pursuer loses sight of the evader after observing it for at least one timestep, maximum steering velocity action is applied in arbitrary direction for 25 timesteps. If the evader is not detected, the pursuer conducts a random walk, changing the input steering velocity at value every \nth{8} timestep, sampled uniformly at random from the normalization interval. The velocity value is controlled to be at maximum forward velocity. For the heuristic evader baselines, we have three different baselines: 1. a random walk algorithm for the point-mass evader, sampling linear acceleration action values from the normalization interval uniformly at random every 25 steps; 2. a greedy evader that moves away from the pursuer upon detection and stands still otherwise; 3. and the Rash evader from Quattrini et al. \cite{quattrini2018search}. To benchmark BiMDN for the second question; both pursuer and evader use the following baselines: 1. vanilla MADDPG 2. image-based latent belief state representation for Gaussian mixtures, inspired by \cite{engin2021learning} 3. simple Unscented Kalman Filter estimator as the belief state with 0.1 noise and initial covariance values as well as sigma points with parameters set as $\alpha=0.1, \beta=2.0, \kappa=1$; motivated by the particle filter method \cite{catellani2023distributed, wan2000unscented}. In a seperate subsection, we benchmark our method against differential game Hamilton-Jacobi solver \cite{bui2022optimizeddp} optimal policies in point-mass vs. point-mass scenario, which is of fewer dimensions and feasible to compute. 
\vspace{-5pt}
\subsection{Multi-Agent Training}\label{Multi-Agent Training}

During the training of our method, both agents start from scratch and improve their performance by playing against each other over 20000 episodes. We set $\Delta t = 0.1\ s$ and maximum number of timesteps as $\mathcal{T}=400$ with 2-frame stacking and skipping. We spawn 8 parallel vectorized environments using the AgileRL Deep Reinforcement Learning Library \cite{Ustaran-Anderegg_AgileRL}. Actor and critic networks are both multilayer perceptrons (MLP) and comprise of 2 hidden layers of size 128 each and ReLU activations. We set replay buffer size at 100000 with 512 batch size, learning rate at 0.0005 for both actor and critic networks with Adam \cite{kingma2014adam} optimizer, exploration noise as Gaussian noise with 0.1 standard deviation and 0 mean. Discount factor ($\gamma$) and soft target update parameter ($\tau$) are set at 0.99 and 0.005, respectively. In our simulation experiments BiMDNs use 1 layer of bidirectional LSTMs, which contain 2 seperate LSTMs for forward and backward passes with 128 hidden units each followed by a fully connected layer of 32 output features. The network thereafter splits into 3 heads with a fully connected layer in each head before the output layers. For cluster weights $\Pi^i_t$, the number of output features is equal to the number of cluster centers in the corresponding layer, other heads have twice the number of output features, representing 2D coordinates. Outputs' activation functions are softmax, Exponential Linear Unit (offset with +1 to enforce nonnegativity) for cluster weights $\Pi^i_t$ and standard deviations $\Sigma^i_t$, respectively. Cluster centers $M^i_t$ are directly predicted by the fully connected layers and number of clusters is empirically set at 3. BiMDN architecture is presented in Fig. \ref{fig:bimdn-details}. 200 previous observations are used for generating the belief states, downsampled by the factor of 10, corresponding to $1\ Hz$. BiMDNs have their own buffers of size 25000 and have a warm-up period of 10000 samples. After that, belief networks are updated every 10 timesteps with samples collected from the corresponding buffers; in batches of 256 samples with Adam optimizer and learning rate of $0.002$.
% \begin{figure}[b]
% \vspace{-15pt}
% \centering
% \subfloat{\includegraphics[width = 1.0\linewidth]{figures/maddpg_belief_loss_npy.pdf}}
% \caption{\textbf{Belief loss regime for both agents:} AAAA}
% \vspace{-15pt}
% \label{fig:belief_loss}
% \end{figure}
The workspace $\mathcal{E}$ is of size $(16,16)$ in meters along the x and y-axes and is centered at $(0,0)$. Starting positions of the agents are sampled uniformly at random within $\mathcal{E}$. Reward coefficients are set as $\lambda_{capture} = 1000$, $\lambda_{d} = 1$, $\lambda_{t} = 1$. We use a simple curriculum learning method which starts training with stronger sensor coverage for the pursuer with $V^p_{\alpha}=2\pi\ rad$  and linearly decays to the training model parameter over $\frac{3}{10}$ of the number of total training episodes. We also start evader velocity at $0\ \frac{m}{s}$ and increase linearly over the same curriculum period until it reaches its defined value. We empirically find that having a curriculum strategy greatly helps the pursuer at the initial training episodes as exploration is guided poorly by random noise, resulting in a very sparse reward scheme.

For training, pursuer dynamic model parameters are set as follows: rear and front axle distances from the center-of-mass $l_f, l_r = 0.15\ m$, steering angle $\theta$ joint limits are $\pm 0.34\ rad$, steering angle velocity $\dot{\theta}$ limits are $\pm 3.2\ \frac{rad}{s}$. Minimum and maximum longitudinal velocities $v$ are $-1\ \frac{m}{s}$ and $2.5\ \frac{m}{s}$, respectively with $\pm 2\ \frac{m}{s^2}$ maximum acceleration $\dot{v}$ value in both directions. Sensor parameters are set as $V^p_{\alpha}=\frac{2\pi}{3}\ rad$  and $V^p_{r}=6\ m$. Evader dynamic model parameters are: minimum and maximum velocities along the axes are $\pm 1.5\ \frac{m}{s}$ with $\pm 9.81\ \frac{m}{s^2}$ maximum acceleration $\dot{v}$ value in both directions.

\begin{figure}[]
\centering
\subfloat{\includegraphics[width = 1.0\linewidth]{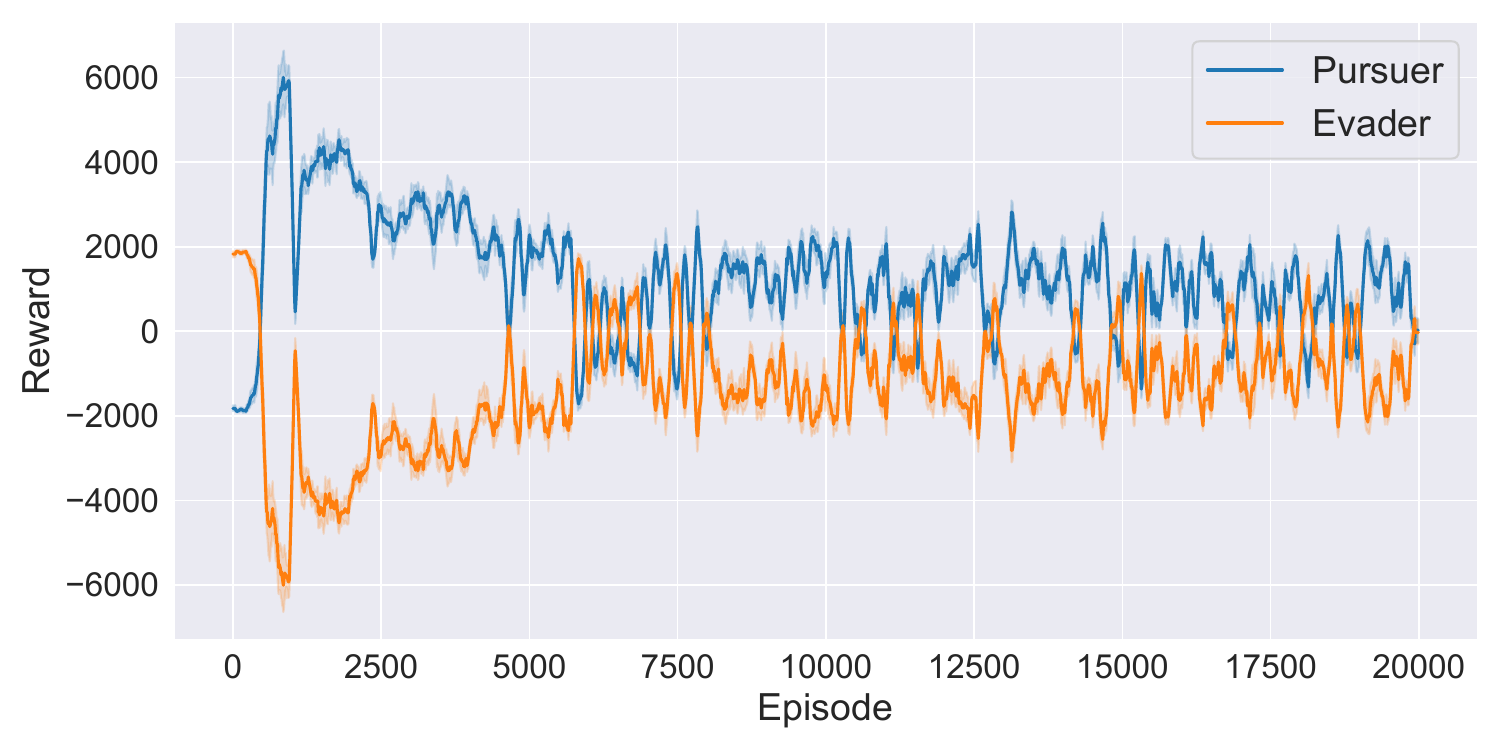}}
\caption{\textbf{Learning regime for both agents:} Pursuer initially starts out by losing almost every episode, after a few hundred episodes the performance catches up and pursuer starts to dominate with the help of curriculum strategy. In response, evader develops its own counter strategies.}
\vspace{-15pt}
\label{fig:training}
\end{figure}

In Fig. \ref{fig:training} we observe a learning regime where the pursuer initially starts out losing almost every episode. Eventually the pursuer performance catches up with the help of the curriculum strategy.  %After that time,  the evader develops its own counter strategies. 
Maximum achievable reward by the pursuer in an individual environment before reset is $\lambda_{capture} = 1000$ (instant capture), yet the rewards are accumulated every 500 timesteps for plotting from 8 parallel environments, therefore each parallel environment can reset multiple times (indicating pursuer domination) before the evaluation timestep, resulting a higher cumulative reward. We propose two variants derived from our BiMDN method, the first one uses weighted mean of Gaussians as the belief state vector during inference time, which represents a pure strategy (Ours), whereas the second variant samples 16 points from the mixture, representing a mixed strategy (Ours-Mixed).

\vspace{-5pt}
\subsection{Performance evaluation against baselines}\label{Performance evaluation against baselines}

In our first set of experiments we investigate the performance of our learned pursuer and evader models against baseline algorithms. For multi-agent training scenario, it is not straightforward to find the best performing algorithm by evaluating algorithms based on total reward or similar metrics at the end of an episode due to the oscillatory nature of the training regime. As a more concrete example, a mediocre pursuer may achieve a deceivingly high score after training against a relatively bad evader or vice versa. 
To mitigate this problem, we create model checkpoints every 250 episodes and evaluate these checkpoints by setting them up against each other (one vs. 32 random samples) after training. The best performing models are selected by evaluating on average capture rates.

%\begin{figure*}
%\centering
%\subfloat[]{\includegraphics[width = 0.33\linewidth]{figures/p_sensor_fov_sweep.pdf}\label{fig:pfov}}
%\subfloat[]{\includegraphics[width = 0.33\linewidth]{figures/sensor_range_sweep.pdf}\label{fig:range}}
%\subfloat[]{\includegraphics[width = 0.33\linewidth]{figures/v_max_sweep.pdf}\label{fig:vmax}}
%\caption{\textbf{Normalized Time-to-Capture values with respect to the pursuer and evader parameters: }Pursuer policies are evaluated against greedy evader policy whereas evader policies are evaluated against learned pursuer policy. Pursuer is able to make good use of its improved sensing capabilities, evader makes rapid gains with velocity until it surpasses the default pursuer velocity.}
%\label{fig:sensor_params}
%\end{figure*}

\begin{table*}[!htb]
\small
\begin{center}
\resizebox{\linewidth}{!}{
\begin{tabular}{lcccccc}
\hline
\textbf{Evader Type}
  & \textbf{Pure Pursuit}
  & \textbf{MADDPG}
  & \textbf{UKF}\cite{catellani2023distributed}
  & \textbf{Image}\cite{engin2021learning}
  & \textbf{Ours}
  & \textbf{Ours-Mixed} \\

\hline

Random Walk
  & 0.59 ($\sigma$=0.36) / 0.67
  & 0.23 ($\sigma$=0.29) / 0.92
  & \textbf{0.27 ($\sigma$=0.22) / 0.98}
  & 0.32 ($\sigma$=0.27) / 0.94
  & \textbf{0.27 ($\sigma$=0.20) / 0.98}
  & \textbf{0.29 ($\sigma$=0.22) / 0.98} \\

Greedy
  & 0.79 ($\sigma$=0.33) / 0.34
  & 0.44 ($\sigma$=0.30) / 0.83
  & 0.31 ($\sigma$=0.21) / 0.94
  & 0.36 ($\sigma$=0.25) / 0.93
  & 0.30 ($\sigma$=0.16) / 0.98
  & \textbf{0.31 ($\sigma$=0.15) / 0.99} \\

Rash \cite{quattrini2018search}
  & 0.74 ($\sigma$=0.37) / 0.39
  & 0.39 ($\sigma$=0.32) / 0.82
  & 0.37 ($\sigma$=0.32) / 0.83
  & 0.40 ($\sigma$=0.34) / 0.82
  & 0.29 ($\sigma$=0.23) / 0.94
  & \textbf{0.32 ($\sigma$=0.20) / 0.99} \\

MADDPG
  & 0.86 ($\sigma$=0.28) / 0.27
  & 0.52 ($\sigma$=0.36) / 0.69
  & 0.42 ($\sigma$=0.32) / 0.79
  & 0.45 ($\sigma$=0.31) / 0.84
  & 0.34 ($\sigma$=0.22) / 0.96
  & \textbf{0.34 ($\sigma$=0.21) / 0.98} \\

UKF\cite{catellani2023distributed}
  & 0.95 ($\sigma$=0.19) / 0.09
  & 0.44 ($\sigma$=0.30) / 0.84
  & 0.49 ($\sigma$=0.36) / 0.70
  & 0.51 ($\sigma$=0.34) / 0.75
  & 0.36 ($\sigma$=0.27) / 0.89
  & \textbf{0.36 ($\sigma$=0.20) / 0.97} \\

Image\cite{engin2021learning}
  & 0.95 ($\sigma$=0.19) / 0.09
  & 0.54 ($\sigma$=0.36) / 0.68
  & 0.46 ($\sigma$=0.32) / 0.82
  & 0.54 ($\sigma$=0.34) / 0.75
  & 0.34 ($\sigma$=0.22) / 0.95
  &  \textbf{0.36 ($\sigma$=0.22) / 0.97} \\

Ours
  & 0.92 ($\sigma$=0.22) / 0.15
  & 0.43 ($\sigma$=0.30) / 0.83
  & 0.52 ($\sigma$=0.35) / 0.70
  & 0.54 ($\sigma$=0.33) / 0.73
  & 0.36 ($\sigma$=0.26) / 0.89
  & \textbf{0.35 ($\sigma$=0.18) / 0.99} \\

Ours-Mixed
  & 0.91 ($\sigma$=0.22) / 0.17
  & 0.51 ($\sigma$=0.35) / 0.69
  & 0.47 ($\sigma$=0.31) / 0.86
  & 0.50 ($\sigma$=0.34) / 0.74
  & \textbf{0.30 ($\sigma$=0.17) / 0.99}
  & \textbf{0.38 ($\sigma$=0.19) / 0.99} \\

\hline
\end{tabular}}
\vspace{-27pt}
\caption{PURSUER CAPTURE TIMES, STANDARD DEVIATIONS / RATES}
\label{tab:timesandrates}
\end{center}
\end{table*}

For learned pursuer and evader strategies, we tested our method against baseline pursuer and evader algorithms as well as against each other over 1500 episodes, using 3 different trained models with 3 random seeds, 500 episodes each. We report the results of the best performing strategies in Table \ref{tab:timesandrates}. The results are normalized between 0 and 1 to represent rates. For any pursuer strategy, a lower capture time and a higher capture rate are better. For any evader strategy, these values are the opposite. From Table \ref{tab:timesandrates} we see that our proposed methods both surpass existing baselines, which indicates that BiMDN improves capture performance in partially observable games. Kalman Filter and image representations also improve over vanilla MADDPG against heuristic evader baselines; however their advantages diminish once they play against learning-based policies, as their pursuer strategies end up being exploitable. The advantage is less pronounced for our trained evader strategies as they struggle to outperform the baseline strategies; showing that belief states are more useful for improving pursuer strategies, although MARL methods are still able to train relatively strong evader policies. To test our method for statistical significance, we use Two Proportion Z-Test over the capture rates' distributions and represent the strongest policies using bold font. Two methods having bold font in the same row means that the p-value is over 0.05 between two distributions.

\vspace{-5pt}

\subsection{Performance evaluation against Hamilton-Jacobi solvers}

In this section we used OptimizedDP \cite{bui2022optimizeddp} to solve a differential game. %The solution yields the gradient of the Value function, which is in turn used to compute the Hamiltonian and the optimal control or disturbance following the $minmax$ or $maxmin$ formulations from the reachability literature. 
Such solvers have a large memory footprint, therefore we reformulate the problem to be a differential game between two homogeneous point-mass agents, reduce the environment size and allowable game duration 
to achieve computational feasibility. The workspace $\mathcal{E}$ is of size $(10,10)$ in meters; maximum number of timesteps is $\mathcal{T}=100$ without frame stacking or skipping. Sensor ranges are $3.75\ m$ and maximum velocities along both axes are increased to $2.5\ \frac{m}{s}$. Our method is trained for 20000 episodes. We follow the same experimental procedure as Section \ref{Performance evaluation against baselines} and report the results in Table \ref{tab:gametimesandrates}. Since optimal control policies are generated for fully observable cases we use BiMDN weighted mean from the Gaussian mixture when the opponent is not observable and compare against the pure strategy accordingly. From Table \ref{tab:gametimesandrates} we see that our pursuer strategies are significantly better than the optimal control policies. Our pursuer method %shows a p-value much smaller than $0.05$ against the evader model as compared to the optimal control method which demonstrates that our proposed method 
improves over existing baselines by a statistically significant margin. The comparative advantage provided by our method to the evader policies is less pronounced with the optimal control pursuer policy being slightly better $(p=0.03)$ against our evader method in terms of capture rate, with the help of the state estimation provided by BiMDN. Results indicate that our proposed method is also competitive with baselines from the reachability literature for simpler problems that are computationally feasible.

\section{Real-World Experiments}

For the real world experiments, a learned pursuer policy was deployed against a human controlled evader. Pursuer policy was directly implemented on an F1TENTH autonomous vehicle~\cite{o2020f1tenth} with the identified system parameters from our previous work \cite{gonultas2023system}. The human-controlled evader was a more nimble autonomous vehicle Nvidia JetRacer~\cite{jetracer}. %Due to real world constraints, the reinforcement learning model state-space representation was slightly modified to use linear velocity and steering angle commands and steering angles are removed from the observation vector (since they are actions). A Gaussian noise with 0 mean and 5\% standard deviation was added to the normalized (between -1 and 1) state and action vectors for robustness.
Sensor ranges were set at $5\ m$ and pursuer sensor field of view angle was set at $\frac{3}{2}\pi\ rad$. The real-world sensor coverage provided by the F1TENTH LiDAR goes up to $20\ m$ at $\frac{3}{2}\pi\ rad$, however we artificially restrict the simulated sensor range at $5\ m$ as our testing environment is significantly smaller at $(10,5.5)$ in meters along the x and y-axes. For 6-DoF real-world state estimation, Phasespace X2E LED motion capture markers were attached to both vehicles and finite differencing was used to calculate velocities at $10\ Hz$. ROS2 was used as the middleware \cite{doi:10.1126/scirobotics.abm6074}. %We demonstrate trained reinforcement learning models are directly transferable to the real system and present sample pursuit evasion trajectories of pursuer intercepting the evader in Fig. \ref{fig:real}. 
For further details on our physical experimental results, please refer to the submitted video attachment. These experiments successfully demonstrate that the learned strategy can be executed on real systems. 
%\begin{figure}[!b]
%\centering
%\subfloat{\includegraphics[width = 0.98\linewidth]{figures/emergence_plot.pdf}}
%\caption{\textbf{Real system trajectories for pursuer intercepting the evader: }We demonstrate trained reinforcement learning models are directly transferable to physical robots. Lighter colors have earlier timestamps, darker colors are later. For further details, please refer to the submitted video attachment.}
%\vspace{-15pt}
%\label{fig:real}
%\end{figure}

\section{Conclusion}\label{Conclusions and Future Directions}
\label{sec:conclusion}

\begin{table}[b]
\small
\begin{center}
\resizebox{\linewidth}{!}{
\begin{tabular}{lcccc}
\hline
\multicolumn{1}{c}{} & \multicolumn{2}{c}{Time-to-Capture}            & \multicolumn{2}{c}{Capture Rate} \\
\textbf{Evader Type} & Optimal Control & \multicolumn{1}{c|}{Ours} & Optimal Control      & Ours      \\ \hline
Optimal Control \cite{bui2022optimizeddp}          & 0.62 ($\sigma$ = 0.40)         & \multicolumn{1}{c|}{0.46 ($\sigma$ = 0.32)}     & 0.47              & 0.83         \\
Ours              & 0.62 ($\sigma$ = 0.40)         & \multicolumn{1}{c|}{0.47 ($\sigma$ = 0.32)}    & 0.50              & 0.80 
\\ \hline
\end{tabular}}
\caption{POINT-MASS CAPTURE TIMES AND RATES}

\label{tab:gametimesandrates}
\end{center}
\vspace{-15pt}
\end{table}
In this paper, we explored a pursuit-evasion game where the pursuer, a car with five state variables, engages with either a similar car or a point mass evader with constraints on the sensing capabilities for both agents. 
We formulated the game as a zero-sum Partially Observable Stochastic Game (POSG) and developed a bidirectional recurrent mixture model named \textit{BiMDN} for complementing the MADDPG algorithm to simultaneously obtain the pursuer and evader policies. By training the players against each other, our approach sidesteps the necessity for pretrained or expert-designed evader policies.
 We also demonstrated that policies obtained in simulation are directly transferable to real autonomous agents and our method is able to handle homogeneous as well as heterogeneous cases in terms of dynamics governing the agents, where existing level-set methods are infeasible. Through this exploration, we show the potential of multi-agent reinforcement learning in navigating complex pursuit-evasion scenarios within dynamic environments. %opening avenues for further research and application within the field of robotics.

There are multiple avenues for future research: in our current formulation, we did not model uncertainty in the players own states. Preliminary experiments indicate that the learned strategies are robust to small errors in state. As the arena gets bigger and if the players do not have perfect global sensors, the uncertainty in state estimation may need to be addressed explicitly. 
Furthermore, if the game takes place in complex environments, the search component of the game becomes prominent. It is unclear if current methods will be able to handle such scenarios in which there will be long sequences with no reward. We will explore these exciting, yet challenging, issues in our future work.

\bibliographystyle{IEEEtran}
\bibliography{reference}

\begin{thebibliography}{10}
\providecommand{\url}[1]{#1}
\csname url@rmstyle\endcsname
\providecommand{\newblock}{\relax}
\providecommand{\bibinfo}[2]{#2}
\providecommand\BIBentrySTDinterwordspacing{\spaceskip=0pt\relax}
\providecommand\BIBentryALTinterwordstretchfactor{4}
\providecommand\BIBentryALTinterwordspacing{\spaceskip=\fontdimen2\font plus
\BIBentryALTinterwordstretchfactor\fontdimen3\font minus
  \fontdimen4\font\relax}
\providecommand\BIBforeignlanguage[2]{{%
\expandafter\ifx\csname l@#1\endcsname\relax
\typeout{** WARNING: IEEEtran.bst: No hyphenation pattern has been}%
\typeout{** loaded for the language `#1'. Using the pattern for}%
\typeout{** the default language instead.}%
\else
\language=\csname l@#1\endcsname
\fi
#2}}

\bibitem{guibas1999visibility}
L.~J. Guibas, J.-C. Latombe, S.~M. LaValle, D.~Lin, and R.~Motwani, ``A
  visibility-based pursuit-evasion problem,'' \emph{International Journal of
  Computational Geometry \& Applications}, vol.~9, no. 04n05, pp. 471--493,
  1999.

\bibitem{isler2005randomized}
V.~Isler, S.~Kannan, and S.~Khanna, ``Randomized pursuit-evasion in a polygonal
  environment,'' \emph{IEEE Transactions on Robotics}, vol.~21, no.~5, pp.
  875--884, 2005.

\bibitem{isaacs1999differential}
R.~Isaacs, \emph{Differential games: a mathematical theory with applications to
  warfare and pursuit, control and optimization}.\hskip 1em plus 0.5em minus
  0.4em\relax Courier Corporation, 1999, originally published in 1965.

\bibitem{bishop1994mixture}
C.~M. Bishop, ``Mixture density networks,'' 1994.

\bibitem{hochreiter1997long}
S.~Hochreiter, ``Long short-term memory,'' \emph{Neural Computation MIT-Press},
  1997.

\bibitem{schuster1999better}
M.~Schuster, ``Better generative models for sequential data problems:
  Bidirectional recurrent mixture density networks,'' \emph{Advances in Neural
  Information Processing Systems}, vol.~12, 1999.

\bibitem{bazzani2016recurrent}
L.~Bazzani, H.~Larochelle, and L.~Torresani, ``Recurrent mixture density
  network for spatiotemporal visual attention,'' \emph{arXiv preprint
  arXiv:1603.08199}, 2016.

\bibitem{lowe2017multi}
R.~Lowe, Y.~I. Wu, A.~Tamar, J.~Harb, O.~Pieter~Abbeel, and I.~Mordatch,
  ``Multi-agent actor-critic for mixed cooperative-competitive environments,''
  \emph{Advances in neural information processing systems}, vol.~30, 2017.

\bibitem{o2020f1tenth}
M.~O'Kelly, H.~Zheng, D.~Karthik, and R.~Mangharam, ``F1tenth: An open-source
  evaluation environment for continuous control and reinforcement learning,''
  \emph{Proceedings of Machine Learning Research}, vol. 123, 2020.

\bibitem{jetracer}
\BIBentryALTinterwordspacing
NVIDIA-AI-IOT, ``{jetracer}.'' [Online]. Available:
  \url{https://github.com/NVIDIA-AI-IOT/jetracer}
\BIBentrySTDinterwordspacing

\bibitem{chung2011search}
T.~H. Chung, G.~A. Hollinger, and V.~Isler, ``Search and pursuit-evasion in
  mobile robotics: A survey,'' \emph{Autonomous robots}, vol.~31, pp. 299--316,
  2011.

\bibitem{exarchos2015suicidal}
I.~Exarchos, P.~Tsiotras, and M.~Pachter, ``On the suicidal pedestrian
  differential game,'' \emph{Dynamic Games and Applications}, vol.~5, pp.
  297--317, 2015.

\bibitem{ruiz2016differential}
U.~Ruiz and R.~Murrieta-Cid, ``A differential pursuit/evasion game of capture
  between an omnidirectional agent and a differential drive robot, and their
  winning roles,'' \emph{International Journal of Control}, vol.~89, no.~11,
  pp. 2169--2184, 2016.

\bibitem{scott2018optimal}
W.~L. Scott and N.~E. Leonard, ``Optimal evasive strategies for multiple
  interacting agents with motion constraints,'' \emph{Automatica}, vol.~94, pp.
  26--34, 2018.

\bibitem{de2021decentralized}
C.~De~Souza, R.~Newbury, A.~Cosgun, P.~Castillo, B.~Vidolov, and D.~Kuli{\'c},
  ``Decentralized multi-agent pursuit using deep reinforcement learning,''
  \emph{IEEE Robotics and Automation Letters}, vol.~6, no.~3, pp. 4552--4559,
  2021.

\bibitem{yang2023large}
H.~Yang, P.~Ge, J.~Cao, Y.~Yang, and Y.~Liu, ``Large scale pursuit-evasion
  under collision avoidance using deep reinforcement learning,'' in \emph{2023
  IEEE/RSJ International Conference on Intelligent Robots and Systems
  (IROS)}.\hskip 1em plus 0.5em minus 0.4em\relax IEEE, 2023, pp. 2232--2239.

\bibitem{mitchell2008flexible}
I.~M. Mitchell, ``The flexible, extensible and efficient toolbox of level set
  methods,'' \emph{Journal of Scientific Computing}, vol.~35, pp. 300--329,
  2008.

\bibitem{helperOC}
\BIBentryALTinterwordspacing
HJReachability, ``helperoc: An optimal control toolbox for hamilton-jacobi
  reachability analysis,'' 2024. [Online]. Available:
  \url{https://github.com/HJReachability/helperOC}
\BIBentrySTDinterwordspacing

\bibitem{bui2022optimizeddp}
M.~Bui, G.~Giovanis, M.~Chen, and A.~Shriraman, ``Optimizeddp: An efficient,
  user-friendly library for optimal control and dynamic programming,''
  \emph{arXiv preprint arXiv:2204.05520}, 2022.

\bibitem{fridovich2020efficient}
D.~Fridovich-Keil, E.~Ratner, L.~Peters, A.~D. Dragan, and C.~J. Tomlin,
  ``Efficient iterative linear-quadratic approximations for nonlinear
  multi-player general-sum differential games,'' in \emph{2020 IEEE
  international conference on robotics and automation (ICRA)}.\hskip 1em plus
  0.5em minus 0.4em\relax IEEE, 2020, pp. 1475--1481.

\bibitem{le2022algames}
S.~Le~Cleac’h, M.~Schwager, and Z.~Manchester, ``Algames: a fast augmented
  lagrangian solver for constrained dynamic games,'' \emph{Autonomous Robots},
  vol.~46, no.~1, pp. 201--215, 2022.

\bibitem{peters2022learning}
L.~Peters, D.~Fridovich-Keil, L.~Ferranti, C.~Stachniss, J.~Alonso-Mora, and
  F.~Laine, ``Learning mixed strategies in trajectory games,'' \emph{arXiv
  preprint arXiv:2205.00291}, 2022.

\bibitem{liu2023learning}
X.~Liu, L.~Peters, and J.~Alonso-Mora, ``Learning to play trajectory games
  against opponents with unknown objectives,'' \emph{IEEE Robotics and
  Automation Letters}, vol.~8, no.~7, pp. 4139--4146, 2023.

\bibitem{zhu2023sequential}
E.~L. Zhu and F.~Borrelli, ``A sequential quadratic programming approach to the
  solution of open-loop generalized nash equilibria,'' in \emph{2023 IEEE
  International Conference on Robotics and Automation (ICRA)}.\hskip 1em plus
  0.5em minus 0.4em\relax IEEE, 2023, pp. 3211--3217.

\bibitem{schwarting2021stochastic}
W.~Schwarting, A.~Pierson, S.~Karaman, and D.~Rus, ``Stochastic dynamic games
  in belief space,'' \emph{IEEE Transactions on Robotics}, vol.~37, no.~6, pp.
  2157--2172, 2021.

\bibitem{suzuki1992searching}
I.~Suzuki and M.~Yamashita, ``Searching for a mobile intruder in a polygonal
  region,'' \emph{SIAM Journal on computing}, vol.~21, no.~5, pp. 863--888,
  1992.

\bibitem{gerkey2006visibility}
B.~P. Gerkey, S.~Thrun, and G.~Gordon, ``Visibility-based pursuit-evasion with
  limited field of view,'' \emph{The International Journal of Robotics
  Research}, vol.~25, no.~4, pp. 299--315, 2006.

\bibitem{stiffler2017persistent}
N.~M. Stiffler, A.~Kolling, and J.~M. O'Kane, ``Persistent pursuit-evasion: The
  case of the preoccupied pursuer,'' in \emph{2017 IEEE International
  Conference on Robotics and Automation (ICRA)}.\hskip 1em plus 0.5em minus
  0.4em\relax IEEE, 2017, pp. 5027--5034.

\bibitem{shishika2018local}
D.~Shishika and V.~Kumar, ``Local-game decomposition for multiplayer
  perimeter-defense problem,'' in \emph{2018 IEEE conference on decision and
  control (CDC)}.\hskip 1em plus 0.5em minus 0.4em\relax IEEE, 2018, pp.
  2093--2100.

\bibitem{noori2014lion}
N.~Noori and V.~Isler, ``The lion and man game on polyhedral surfaces with
  boundary,'' in \emph{2014 IEEE/RSJ International Conference on Intelligent
  Robots and Systems}.\hskip 1em plus 0.5em minus 0.4em\relax IEEE, 2014, pp.
  1769--1774.

\bibitem{engin2021learning}
S.~Engin, Q.~Jiang, and V.~Isler, ``Learning to play pursuit-evasion with
  visibility constraints,'' in \emph{2021 IEEE/RSJ International Conference on
  Intelligent Robots and Systems (IROS)}.\hskip 1em plus 0.5em minus
  0.4em\relax IEEE, 2021, pp. 3858--3863.

\bibitem{zhang2022game}
R.~Zhang, Q.~Zong, X.~Zhang, L.~Dou, and B.~Tian, ``Game of drones: Multi-uav
  pursuit-evasion game with online motion planning by deep reinforcement
  learning,'' \emph{IEEE Transactions on Neural Networks and Learning Systems},
  vol.~34, no.~10, pp. 7900--7909, 2022.

\bibitem{bajcsy2023learning}
A.~Bajcsy, A.~Loquercio, A.~Kumar, and J.~Malik, ``Learning vision-based
  pursuit-evasion robot policies,'' \emph{arXiv preprint arXiv:2308.16185},
  2023.

\bibitem{shapley1953stochastic}
L.~S. Shapley, ``Stochastic games,'' \emph{Proceedings of the national academy
  of sciences}, vol.~39, no.~10, pp. 1095--1100, 1953.

\bibitem{littman1994markov}
M.~L. Littman, ``Markov games as a framework for multi-agent reinforcement
  learning,'' in \emph{Machine learning proceedings 1994}.\hskip 1em plus 0.5em
  minus 0.4em\relax Elsevier, 1994, pp. 157--163.

\bibitem{mitchell2005time}
I.~M. Mitchell, A.~M. Bayen, and C.~J. Tomlin, ``A time-dependent
  hamilton-jacobi formulation of reachable sets for continuous dynamic games,''
  \emph{IEEE Transactions on automatic control}, vol.~50, no.~7, pp. 947--957,
  2005.

\bibitem{zhang2021multi}
K.~Zhang, Z.~Yang, and T.~Ba{\c{s}}ar, ``Multi-agent reinforcement learning: A
  selective overview of theories and algorithms,'' \emph{Handbook of
  reinforcement learning and control}, pp. 321--384, 2021.

\bibitem{pinto2017robust}
L.~Pinto, J.~Davidson, R.~Sukthankar, and A.~Gupta, ``Robust adversarial
  reinforcement learning,'' in \emph{International Conference on Machine
  Learning}.\hskip 1em plus 0.5em minus 0.4em\relax PMLR, 2017, pp. 2817--2826.

\bibitem{bucsoniu2010multi}
L.~Bu{\c{s}}oniu, R.~Babu{\v{s}}ka, and B.~De~Schutter, ``Multi-agent
  reinforcement learning: An overview,'' \emph{Innovations in multi-agent
  systems and applications-1}, pp. 183--221, 2010.

\bibitem{ackermann2019reducing}
J.~Ackermann, V.~Gabler, T.~Osa, and M.~Sugiyama, ``Reducing overestimation
  bias in multi-agent domains using double centralized critics,'' \emph{arXiv
  preprint arXiv:1910.01465}, 2019.

\bibitem{de2020independent}
C.~S. de~Witt, T.~Gupta, D.~Makoviichuk, V.~Makoviychuk, P.~H. Torr, M.~Sun,
  and S.~Whiteson, ``Is independent learning all you need in the starcraft
  multi-agent challenge?'' \emph{arXiv preprint arXiv:2011.09533}, 2020.

\bibitem{rashid2020monotonic}
T.~Rashid, M.~Samvelyan, C.~S. De~Witt, G.~Farquhar, J.~Foerster, and
  S.~Whiteson, ``Monotonic value function factorisation for deep multi-agent
  reinforcement learning,'' \emph{The Journal of Machine Learning Research},
  vol.~21, no.~1, pp. 7234--7284, 2020.

\bibitem{vinyals2019grandmaster}
O.~Vinyals, I.~Babuschkin, W.~M. Czarnecki, M.~Mathieu, A.~Dudzik, J.~Chung,
  D.~H. Choi, R.~Powell, T.~Ewalds, P.~Georgiev, \emph{et~al.}, ``Grandmaster
  level in starcraft ii using multi-agent reinforcement learning,''
  \emph{Nature}, vol. 575, no. 7782, pp. 350--354, 2019.

\bibitem{berner2019dota}
C.~Berner, G.~Brockman, B.~Chan, V.~Cheung, P.~Debiak, C.~Dennison, D.~Farhi,
  Q.~Fischer, S.~Hashme, C.~Hesse, \emph{et~al.}, ``Dota 2 with large scale
  deep reinforcement learning,'' \emph{arXiv preprint arXiv:1912.06680}, 2019.

\bibitem{silver2018general}
D.~Silver, T.~Hubert, J.~Schrittwieser, I.~Antonoglou, M.~Lai, A.~Guez,
  M.~Lanctot, L.~Sifre, D.~Kumaran, T.~Graepel, \emph{et~al.}, ``A general
  reinforcement learning algorithm that masters chess, shogi, and go through
  self-play,'' \emph{Science}, vol. 362, no. 6419, pp. 1140--1144, 2018.

\bibitem{silver2016mastering}
D.~Silver, A.~Huang, C.~J. Maddison, A.~Guez, L.~Sifre, G.~Van Den~Driessche,
  J.~Schrittwieser, I.~Antonoglou, V.~Panneershelvam, M.~Lanctot,
  \emph{et~al.}, ``Mastering the game of go with deep neural networks and tree
  search,'' \emph{nature}, vol. 529, no. 7587, pp. 484--489, 2016.

\bibitem{brunnbauer2022latent}
A.~Brunnbauer, L.~Berducci, A.~Brandst{\'a}tter, M.~Lechner, R.~Hasani, D.~Rus,
  and R.~Grosu, ``Latent imagination facilitates zero-shot transfer in
  autonomous racing,'' in \emph{2022 international conference on robotics and
  automation (ICRA)}.\hskip 1em plus 0.5em minus 0.4em\relax IEEE, 2022, pp.
  7513--7520.

\bibitem{monahan1982state}
G.~E. Monahan, ``State of the art—a survey of partially observable markov
  decision processes: theory, models, and algorithms,'' \emph{Management
  science}, vol.~28, no.~1, pp. 1--16, 1982.

\bibitem{kaelbling1998planning}
L.~P. Kaelbling, M.~L. Littman, and A.~R. Cassandra, ``Planning and acting in
  partially observable stochastic domains,'' \emph{Artificial intelligence},
  vol. 101, no. 1-2, pp. 99--134, 1998.

\bibitem{silver2010monte}
D.~Silver and J.~Veness, ``Monte-carlo planning in large pomdps,''
  \emph{Advances in neural information processing systems}, vol.~23, 2010.

\bibitem{wayne2018unsupervised}
G.~Wayne, C.-C. Hung, D.~Amos, M.~Mirza, A.~Ahuja, A.~Grabska-Barwinska,
  J.~Rae, P.~Mirowski, J.~Z. Leibo, A.~Santoro, \emph{et~al.}, ``Unsupervised
  predictive memory in a goal-directed agent,'' \emph{arXiv preprint
  arXiv:1803.10760}, 2018.

\bibitem{han2019variational}
D.~Han, K.~Doya, and J.~Tani, ``Variational recurrent models for solving
  partially observable control tasks,'' \emph{arXiv preprint arXiv:1912.10703},
  2019.

\bibitem{moreno2021neural}
P.~Moreno, E.~Hughes, K.~R. McKee, B.~A. Pires, and T.~Weber, ``Neural
  recursive belief states in multi-agent reinforcement learning,'' \emph{arXiv
  preprint arXiv:2102.02274}, 2021.

\bibitem{wang2023learning}
A.~Wang, A.~C. Li, T.~Q. Klassen, R.~T. Icarte, and S.~A. McIlraith, ``Learning
  belief representations for partially observable deep rl,'' in
  \emph{International Conference on Machine Learning}.\hskip 1em plus 0.5em
  minus 0.4em\relax PMLR, 2023, pp. 35\,970--35\,988.

\bibitem{althoff2017commonroad}
M.~Althoff, M.~Koschi, and S.~Manzinger, ``Commonroad: Composable benchmarks
  for motion planning on roads,'' in \emph{2017 IEEE Intelligent Vehicles
  Symposium (IV)}.\hskip 1em plus 0.5em minus 0.4em\relax IEEE, 2017, pp.
  719--726.

\bibitem{coulter1992implementation}
R.~C. Coulter \emph{et~al.}, \emph{Implementation of the pure pursuit path
  tracking algorithm}.\hskip 1em plus 0.5em minus 0.4em\relax Carnegie Mellon
  University, The Robotics Institute, 1992.

\bibitem{quattrini2018search}
A.~Quattrini~Li, R.~Fioratto, F.~Amigoni, and V.~Isler, ``A search-based
  approach to solve pursuit-evasion games with limited visibility in polygonal
  environments,'' in \emph{Proceedings of the 17th International Conference on
  Autonomous Agents and MultiAgent Systems}, 2018, pp. 1693--1701.

\bibitem{catellani2023distributed}
M.~Catellani and L.~Sabattini, ``Distributed control of a limited angular
  field-of-view multi-robot system in communication-denied scenarios: A
  probabilistic approach,'' \emph{IEEE Robotics and Automation Letters},
  vol.~9, no.~1, pp. 739--746, 2023.

\bibitem{wan2000unscented}
E.~A. Wan and R.~Van Der~Merwe, ``The unscented kalman filter for nonlinear
  estimation,'' in \emph{Proceedings of the IEEE 2000 adaptive systems for
  signal processing, communications, and control symposium (Cat. No.
  00EX373)}.\hskip 1em plus 0.5em minus 0.4em\relax Ieee, 2000, pp. 153--158.

\bibitem{Ustaran-Anderegg_AgileRL}
\BIBentryALTinterwordspacing
N.~Ustaran-Anderegg and M.~Pratt, ``{AgileRL}.'' [Online]. Available:
  \url{https://github.com/AgileRL/AgileRL}
\BIBentrySTDinterwordspacing

\bibitem{kingma2014adam}
D.~P. Kingma, ``Adam: A method for stochastic optimization,'' \emph{arXiv
  preprint arXiv:1412.6980}, 2014.

\bibitem{gonultas2023system}
B.~M. Gonultas, P.~Mukherjee, O.~G. Poyrazoglu, and V.~Isler, ``System
  identification and control of front-steered ackermann vehicles through
  differentiable physics,'' in \emph{2023 IEEE/RSJ International Conference on
  Intelligent Robots and Systems (IROS)}.\hskip 1em plus 0.5em minus
  0.4em\relax IEEE, 2023, pp. 4347--4353.

\bibitem{doi:10.1126/scirobotics.abm6074}
S.~Macenski, T.~Foote, B.~Gerkey, C.~Lalancette, and W.~Woodall, ``Robot
  operating system 2: Design, architecture, and uses in the wild,''
  \emph{Science Robotics}, vol.~7, no.~66, p. eabm6074, 2022.

\end{thebibliography}

%\clearpage
%\appendix
%\renewcommand{\thesection}{\Alph{section}.\arabic{section}}
%\setcounter{section}{0}
% \input{8-appendix}
\end{document}